\documentclass[10pt,twocolumn,letterpaper]{article}

\usepackage[pagenumbers]{cvpr} 

\usepackage[ruled,noend]{algorithm2e} 

\SetAlFnt{\small}
\SetAlCapFnt{\small}
\SetAlCapNameFnt{\small}
\SetAlCapHSkip{0pt}
\IncMargin{-\parindent}
\usepackage{ifpdf}
\usepackage{graphicx}

\usepackage{color}
\definecolor{cyan}{cmyk}{1,0,0,0}
\definecolor{darkgreen}{rgb}{0,0.5,0}
\definecolor{orange}{rgb}{1,0.5,0}
\definecolor{magenta}{cmyk}{0,1,0,0}
\definecolor{darkyellow}{cmyk}{0,0,0.75,0}
\definecolor{gray}{rgb}{0.8,0.8,0.8}

\usepackage{amssymb} 
\usepackage{amsmath} 

\usepackage[toc,page]{appendix} 
\usepackage{wrapfig} 
\usepackage{comment}
\usepackage{bm}
\usepackage{cuted} 
\usepackage{lipsum} 
\usepackage{mathtools} 
\usepackage{algorithmicx}
\usepackage{algpseudocode}
\usepackage{nicefrac}

\usepackage{gensymb}
\usepackage{float}
\usepackage{soul}
\usepackage{array}
\usepackage{multirow}
\usepackage{hhline}
\usepackage{setspace}
\usepackage{nicefrac}

\algdef{SE}[DOWHILE]{Do}{doWhile}{\algorithmicdo}[1]{\algorithmicwhile\ #1}%

\makeatletter
\renewcommand{\ALG@beginalgorithmic}{\small}
\makeatother

\newcommand{\DELETE}[1]{} 
\newcommand{\IGNORE}[1]{}

\usepackage{graphicx}
\usepackage{amsmath}
\usepackage{amssymb}
\usepackage{booktabs}

\usepackage[pagebackref,breaklinks,colorlinks]{hyperref}

\def\cvprPaperID{6491} 
\def\confName{CVPR}
\def\confYear{2022}

\begin{document}
	
\title{%
All-photon Polarimetric Time-of-Flight Imaging
} 
\author{
 \hspace{-10mm}
Seung-Hwan Baek$^{*}$
\qquad
Felix Heide  \vspace{2mm}\\
	Princeton University
}



\twocolumn[{%
\vspace{-0mm}
\renewcommand\twocolumn[1][]{#1}%
\maketitle
\vspace{-0mm}
\thispagestyle{empty}
}]
\newcommand\blfootnote[1]{%
  \begingroup
  \renewcommand\thefootnote{}\footnote{#1}%
  \addtocounter{footnote}{-1}%
  \endgroup
}
 \blfootnote{[$^*$] Now at POSTECH.}
\begin{abstract}
Time-of-flight (ToF) sensors provide an image modality fueling diverse applications, including LiDAR in autonomous driving, robotics, and augmented reality. Conventional ToF imaging methods estimate the depth by sending pulses of light into a scene and measuring the ToF of the first-arriving photons directly reflected from a scene surface without any temporal delay. As such, all photons following this first response are typically considered as unwanted noise. 
In this paper, we depart from the principle of using first-arriving photons and propose an all-photon ToF imaging method by incorporating the temporal-polarimetric analysis of first- and late-arriving photons, which possess rich scene information about its geometry and material. To this end, we propose a novel temporal-polarimetric reflectance model, an efficient capture method, and a reconstruction method that exploits the temporal-polarimetric changes of light reflected by the surface and sub-surface reflection.
The proposed all-photon polarimetric ToF imaging method allows for acquiring depth, surface normals, and material parameters of a scene by utilizing all photons captured by the system, whereas conventional ToF imaging only obtains coarse depth from the first-arriving photons.
We validate our method in simulation and experimentally with a prototype.
\end{abstract} 

\begin{figure}[t]
  \centering
  \includegraphics[width=\linewidth]{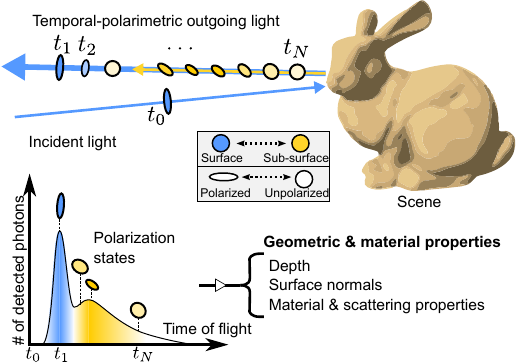}
  \caption{\label{fig:teaser}%
  All-photon polarimetric ToF imaging:
  In contrast to the conventional ToF imaging systems analyzing the first-reflected surface reflection, we propose to capture and exploit the temporal-polarimetric characteristics of both surface and sub-surface reflections.
  This enables us to infer scene parameters (depth, normals, and material parameters) from all captured photons.
  }
\end{figure}

\section{Introduction}
\label{sec:introduction}
Depth cameras have equipped us with the ability to acquire scene geometry that is essential for 3D vision both \emph{directly} as a geometric input to a perception or reconstruction method, and \emph{indirectly}, as supervised training data in rich 3D vision datasets~\cite{replica19arxiv,dai2017scannet,chang2017matterport3d}. As such, depth sensor have fueled diverse applications across domains, including autonomous vehicles, virtual and augmented reality, mobile photography, and robotics. A large body of prior work explores different approaches to extract depth, including parallax~\cite{scharstein2002taxonomy}, defocus~\cite{levin2007image}, double refraction~\cite{baek2016birefractive}, and correlation imaging~\cite{lange2001solid}.
In particular, \textit{pulsed} ToF cameras~\cite{renker2006geiger} have been rapidly adopted across application domains for their high depth precision, low baseline, long range and robustness to scene texture.

ToF cameras directly measure the round-trip return time of light emitted into a scene. This requires high-speed sensors and electronics that can time-tag photon arrival events, such as single-photon avalanche diode (SPAD) and time-correlated single-photon counting (TSCPC) electronics. By repeating pulses and recording arrival events hundreds of thousands of times per measurement, these pulsed ToF cameras are capable of recording an entire temporal histogram for each pixel, describing the number of photons detected by the sensor for each travel time bin.

Photons that arrive at different times at a single pixel can be broadly categorized into two temporally distinct components that sum up to the total measured photons.
The first group of photons originates from the light directly reflected off from an object surface.
Therefore, this surface reflection allows for estimating the object depth directly from its ToF measurement. Surface reflection often has higher intensity in comparison to the other component. As a result, one common practice for ToF depth imaging is to estimate the ToF through peak-finding over the photon counts. In contrast to the surface reflection, the second group of photons comes from the light that penetrates the object surface, undergoes sub-surface scattering, and comes out of the surface to the detector.
This sub-surface scattered light arrives at the sensor with a longer travel time than the light originating from surface reflections.
\emph{At first glance, sub-surface scattering appears to harm accurate depth estimation} as it disturbs the true ToF by an additional unknown scattering time.
Existing ToF methods typically discard this sub-surface observation and only rely on the surface reflection.

In this work, we treat this sub-surface scattering component as a signal instead of noise in estimating geometric and material properties of a scene.
The proposed \emph{all-photon ToF imaging method combines ToF imaging with polarization analysis}, exploiting the unique temporal-polarimetric signature of surface and sub-surface reflections, as shown in Figure~\ref{fig:teaser}. 
The proposed method hinges on a novel temporal-polarimetric bidirectional reflectance distribution function (BRDF) model that describes both surface and sub-surface reflections in temporal and polarimetric domains.
We implement this BRDF model with differentiable operators allowing us to pose \emph{scene reconstruction as an inverse rendering problem.}
In order to reduce the acquisition burden of the high-dimensional polarimetric-temporal data, we also propose a light-efficient temporal-polarimetric acquisition approach that captures the temporal-polarimetric signals of a scene combined with a learned optical ellipsometry.
We demonstrate that all-photon polarimetric ToF imaging can record diverse scene information, including depth, normals, and material parameters from temporal-polarimetric measurements -- both in simulation and experimentally.
Specifically, we make the following contributions.
\begin{itemize}
  \item We devise an all-photon polarimetric ToF imaging method that combines ToF imaging and polarization analysis to jointly exploit surface and sub-surface reflected photons for scene reconstruction.
  \item We present the first temporal-polarimetric BRDF model which characterizes surface and sub-surface reflection based on micro-facet theory and Stokes-Mueller formalism.
  \item We propose an efficient temporal-polarimetric acquisition approach tailored for polarimetric imaging with polarized-laser illumination.
  \item We validate our method in simulation and with an experimental prototype for capturing depth, normals, and material parameters.
\end{itemize}

\paragraph{Limitations}
As a result of the mechanical rotation of the polarizing optics and sequential scanning of temporal histograms, the acquisition of the temporal-polarimetric data takes five hours for each scene. We note that the limited number of temporal-polarimetric photons and low sensor quantum efficiency makes the acquisition challenging. In the future, accelerating the polarimetric capture with electronically-controllable liquid crystal modulators instead of mechanically rotating polarizers, or using a polarization array filters placed on photon-efficient SPAD sensor arrays, may eliminate the need for sequential acquisition. 
\section{Related Work}
\label{sec:relatedwork}


\paragraph{ToF Imaging}
Modern ToF imaging methods can be broadly categorized into correlation and pulsed approaches. Correlation ToF methods emit intensity-modulated light into a scene and \emph{indirectly} estimate the round-trip time of the light returned back to the sensor by performing intensity demodulation~\cite{lange2001solid,zhang2012microsoft}. 
In contrast, pulsed ToF methods \emph{directly} measure ToF of light in an emitted pulse. These methods combine picosecond-pulsed illumination with time-resolved sensors, such as a SPAD, synchronized with the laser using TSCPC electronics~\cite{mccarthy2009long,heide2018sub}. Not only have pulsed ToF approaches been deployed broadly for LiDAR systems in autonomous vehicles~\cite{schwarz2010mapping}, robotics~\cite{kim2021nanophotonics}, and space missions~\cite{abdalati2010icesat}, but recently in consumer mobile phones such as the iPhone 12 Pro. Typical LiDAR sensors extract depth by reporting a detection event for the \emph{first-arriving} photons among all photons reflected from a scene surface. To this end, existing methods implement peak finding via analog thresholding~\cite{schwarz2010mapping}, or peak-finding via algorithms~\cite{shin2015photon}, spatial coherence~\cite{rapp2017few}, temporal coherence~\cite{shin2016photon}, and learned priors~\cite{lindell2018single}. This acquisition principle also means that \emph{photons arriving later are ignored}, along with rich material and geometric information that these photons carry about the scene.
Researchers have attempted to utilize this hidden information by modeling a more comprehensive \emph{temporal} response of a scene with analytic temporal BRDF models~\cite{wu2014decomposing,satat2016all,satat2018towards}.
However, temporal-only analysis restricts them to only extracting scene depth from the \emph{fitted} first-arriving photons, while the later arriving photons are only used for obtaining a better fit of the first-arriving photons.
In contrast, we analyze \emph{all} detected photons by jointly analyzing their polarization and temporal characteristics to extract geometric and material properties of a scene.

\paragraph{Polarimetric Imaging}
A rich body of work has explored using polarization of light for various visual-computing tasks. Early works rely on cross-polarization, which equips illumination and detector with orthogonal polarization filters. This allows for separating polarized reflection from unpolarized reflection, or vice versa~\cite{schechner2001instant,nayar1997separation}.
This principle has led to many applications such as seeing through scattering media~\cite{treibitz2008active} and diffuse-specular separation~\cite{ghosh2010circularly,ma2007rapid}.
Recently, polarization cameras with spatially multiplexed linear polarizers have been used for the acquisition of surface normals~\cite{atkinson2006recovery,kadambi2015polarized}. These method explicitly or implicitly rely on polarimetric BRDFs that couple local geometry to a polarization response~\cite{baek2018simultaneous,baek2020image}.
Combining polarization and ToF imaging has largely been an unexplored field.
Callenberg et al.~\cite{callenberg2017snapshot} investigated polarization-difference imaging by configuring a two-illumination correlation ToF camera in a cross-polarization mode.
Most relevant to our work, Baek et al.~\cite{baek2021probe} combine polarization imaging and ToF imaging to capture the temporal-polarimetric response of a scene. Their method is \emph{unable to extract geometry and reflectance} from the measurements due to the absence of a temporal-polarimetric BRDF model that links the scene parameters to the observation. We close this gap in this work and introduce \emph{the first temporal-polarimetric BRDF model}, which makes it possible to estimate scene parameters directly from temporal-polarimetric observations. 

\paragraph{Polarization LiDAR}
Interestingly, combining polarization with ToF imaging has been studied for decades in geoscience~\cite{sassen2005polarization}.
Notably, researchers in cloud science have applied polarization analysis to LiDAR systems in order to examine the cloud properties and its aerosol concentration by sending out polarized laser pulses to the cloud and measure how much of returned light, at a given travel time, is depolarized~\cite{sassen1991polarization,winker2009overview,goldstein2017polarized}. This principle, first studied by Schotland et al.~\cite{schotland1971observations}, has led to remarkable success, not only in cloud analysis but also in other fields like biology~\cite{huffman2020real} and ocean science~\cite{vasilkov2001airborne}. The proposed method shares this motivation in that we combine polarization and ToF imaging, however, we tackle the problem of scene reconstruction.


\begin{figure}[t]
  \centering
  \includegraphics[width=1\linewidth]{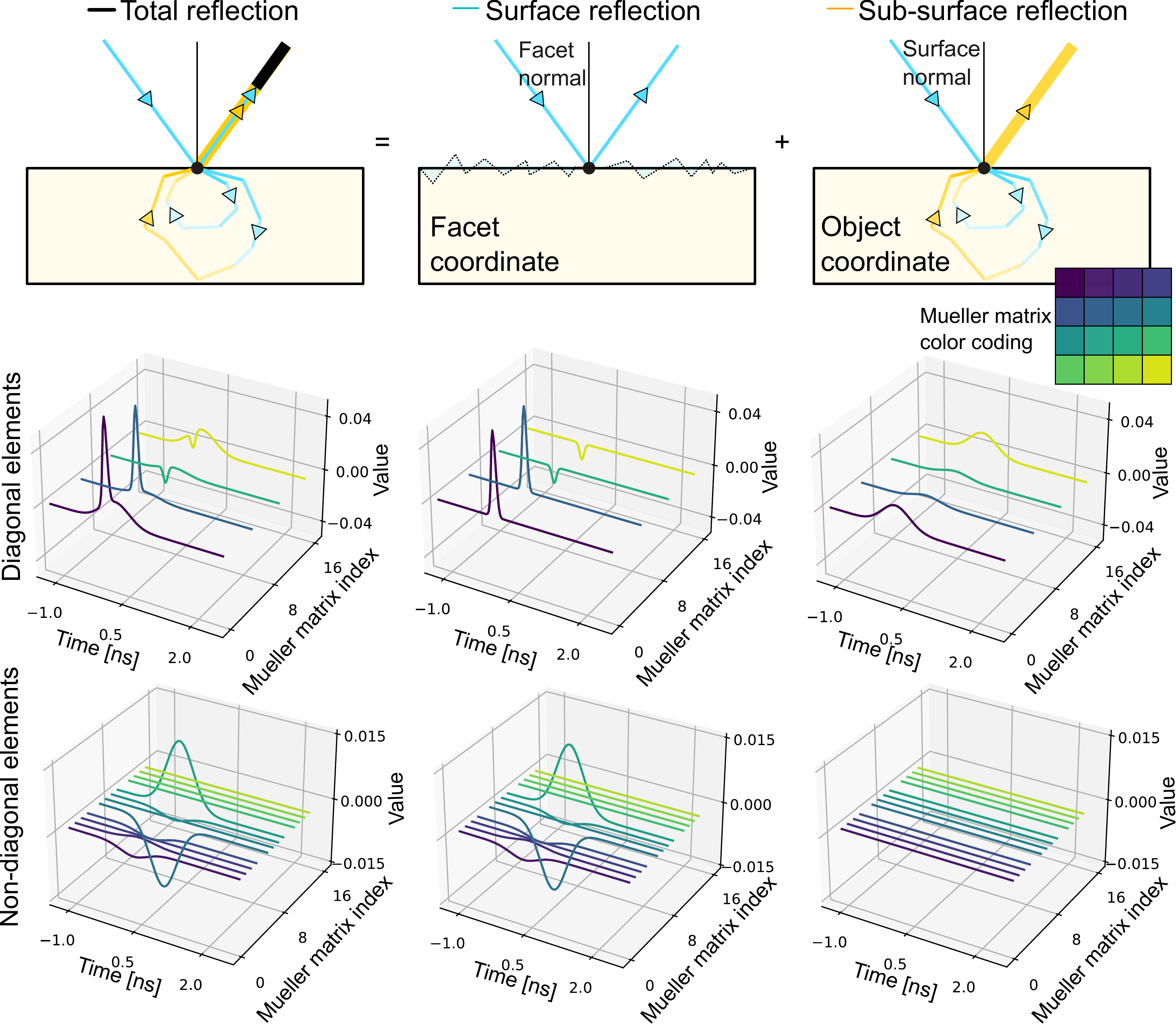}
  \caption{\label{fig:brdf}%
  We introduce the first temporal-polarimetric reflectance model which describes the temporal-polarization change of surface and sub-surface reflection. We show an example temporal-polarimetric reflectance simulated using our model.
  }
\end{figure}

\section{Temporal-polarimetric Light Transport}
\label{sec:image_formation}
The proposed method hinges on a novel analytical model for temporal-polarimetric light transport, which we introduce in this section.
Our model is the first to describe both surface and sub-surface reflection based on the Stokes-Mueller formalism. We represent polarized light and reflectance with a Stokes vector and a Mueller matrix~\cite{collett2005field}.

\subsection{Temporal-polarimetric Reflectance}
\label{sec:brdf}
Next, we introduce a temporal-polarimetric reflectance model $\mathbf{M}(\tau, \boldsymbol{\omega}_i, \boldsymbol{\omega}_o)$. This model describes how light polarization and intensity change when impinging on a surface with given incident and outgoing direction of light ($\boldsymbol{\omega}_i$ and $\boldsymbol{\omega}_o$), and with temporal delay upon the arrival of incident light ($\tau$). Specifically, we model reflectance $\mathbf{M}$ as a sum of surface and sub-surface reflection ($\mathbf{M}_s$ and $\mathbf{M}_{ss}$)
\begin{equation}\label{eq:tpBRDF}
\mathbf{M}\left(\tau, \boldsymbol{\omega}_i, \boldsymbol{\omega}_o\right) = \mathbf{M}_s\left(\tau, \boldsymbol{\omega}_i, \boldsymbol{\omega}_o\right) + \mathbf{M}_{ss}\left(\tau, \boldsymbol{\omega}_i, \boldsymbol{\omega}_o\right).
\end{equation}
We describe both components in the following, see also Figure~\ref{fig:brdf}.

\paragraph{Surface Reflection}
Some of the incident photons are immediately scattered in a surface reflection, including both retro-reflection and specular reflection~\cite{oren1994generalization}.
Following microfacet theory~\cite{torrance1967theory}, we can model a surface as a collection of microfacets. Surface reflection originates from the microfacets oriented orthogonal to the halfway vector $\mathbf{h}=\frac{\boldsymbol{\omega}_i+\boldsymbol{\omega}_o}{2}$. To describe the polarimetric change induced by this microfacet reflection, existing work has relied on the Fresnel-reflection Mueller matrix $\mathbf{F}_R$. This is a function of the refractive index of the material $\eta$ ~\cite{collett2005field,baek2018simultaneous,hyde2009geometrical}. In our work, we also consider time in addition to polarization, which can induce a slight temporal delay of the surface reflection, e.g., originating from reflection between near surfaces or microfacets. This results in the depolarization of light, which we model with a time-varying attenuation Mueller matrix $\mathbf{D}^{s}(\tau)$ using polarization-dependent Gaussian distributions, that is
$\mathbf{D}^s_{i}(\tau) = a^s_{i} \mathrm{exp}({ \frac{(\tau-\mu^s_i)^2}{2(\sigma^s_i)^2}})$,
where $\mathbf{D}^s_{i}$ is the $i$-th diagonal element, $a^s_i$ is the amplitude, $\mu^s_i$ and $(\sigma^s_i)^2$ are the mean and variance.
The matrix $\mathbf{D}^s$ can attenuate the intensity and polarization components of incident light at different degrees. Combining this with Smith's shading and masking function~\cite{heitz2014understanding} and the GGX normal distribution function~\cite{walter2007microfacet}, we arrive at the proposed temporal-polarimetric surface-reflection model
\begin{equation}\label{eq:tpBRDF_s}
\mathbf{M}_s\left(\tau, \boldsymbol{\omega}_i, \boldsymbol{\omega}_o\right) = \frac{D(\theta_h; \sigma)G(\theta_i, \theta_o; m)}{4\cos\theta_i \cos\theta_o} \mathbf{D}^s(\tau) \mathbf{F}_R,
\end{equation}
where $\theta_h=\cos^{-1}(\mathbf{h}\cdot\mathbf{n})$ is the halfway angle, $\mathbf{n}$ is the surface normals, $m$ is the surface roughness, $\theta_i=\cos^{-1}\mathbf{n}\cdot\boldsymbol{\omega}_i$ and $\theta_o=\cos^{-1}\mathbf{n}\cdot\boldsymbol{\omega}_o$ are the incident/outgoing angles.

\paragraph{Sub-surface Reflection}
The second type of reflection originates from the light transmitted into the object. The transmitted light undergoes sub-surface scattering and eventually leaves the object. The Fresnel-transmission Mueller matrix $\mathbf{F}_T$~\cite{collett2005field} describes the polarization change of light passing through the interface between the two mediums (air and bulk material). Baek et al.~\cite{baek2018simultaneous} combined this Fresnel matrix with a perfect depolarization Mueller matrix to describe sub-surface reflection with an assumption that light becomes completely unpolarized after sub-surface scattering. However, in reality, depolarization by sub-surface scattering is highly correlated with the path length of light; sub-surface scattering with longer path lengths results in more depolarization~\cite{germer2020evolution}. Accurately modeling this behavior is essential for the proposed temporal-polarimetric imaging method because path length decides the temporal delay. To this end, we model sub-surface scattering with time-varying depolarization. We employ time-varying Gaussian distributions as the diagonal elements of a depolarization Mueller matrix $\mathbf{D}^{ss}(\tau)$ similar to the surface-reflection model with the parameters $a_i^{ss}$, $\mu_i^{ss}$, $\sigma_i^{ss}$. 

For sub-surface scattering, we also consider coordinate conversion for Stokes vectors. Specifically, for incident/outgoing light to/from a surface, we define the corresponding Stokes vector in a coordinate that has a $z$ axis matching the light propagation direction, while the $x$ and $y$ axes can be arbitrarily chosen to form an orthonormal basis~\cite{collett2005field}.
Among infinite candidates for the $x$ and $y$ axes, we choose them following the halfway coordinate approach~\cite{baek2020image}.
In halfway coordinates, we define the $y$ axis on a plane spanning the halfway vector $\mathbf{h}$ and the propagation direction, which determines $x$ axis as $y \times z$.
In this convention, we convert the halfway coordinate of the incident/outgoing light to a surface normal coordinate. This results in the $y$ axis to fall in a plane that spans the surface normal and the propagation direction because the transmission event itself is defined on the corresponding surface normal coordinate~\cite{baek2018simultaneous}.

With all components described above, we define our temporal-polarimetric sub-surface reflection model
\begin{equation}\label{eq:tpBRDF_d}
\mathbf{M}_{ss}\left(\tau,\boldsymbol{\omega}_i, \boldsymbol{\omega}_o\right) = \mathbf{C}_{\mathrm{n}\rightarrow\mathrm{o}}\mathbf{F}_T^o\mathbf{D}^{ss}(\tau)\mathbf{F}_T^i\mathbf{C}_{\mathrm{i}\rightarrow\mathrm{n}},
\end{equation}
where $\mathbf{C}_{\mathrm{i}\rightarrow\mathrm{n}}$ and $\mathbf{C}_{\mathrm{n}\rightarrow\mathrm{o}}$ are the coordinate-conversion Mueller matrices from the halfway coordinates of incident and outgoing light to the surface normal coordinate~\cite{collett2005field}. Here, $\mathbf{F}_T^i$ and $\mathbf{F}_T^o$ are the Fresnel transmission Mueller matrices for the incident and outgoing light which both depend on the refractive index of the material $\eta$.
Note that our surface-reflection model does not require coordinate conversion, as surface reflection is already defined in halfway coordinates following microfacet theory~\cite{torrance1967theory}.

\subsection{Temporal-polarimetric Rendering}
The proposed temporal-polarimetric reflectance model returns a Mueller matrix $\mathbf{M}(\tau,\boldsymbol{\omega}_i, \boldsymbol{\omega}_o)$ at a temporal delay $\tau$ and incident and outgoing directions, $\boldsymbol{\omega}_i$ and $\boldsymbol{\omega}_o$.
Incorporating this Mueller matrix into the rendering equation~\cite{kajiya1986rendering} describes the Stokes vector of the outgoing light at each temporal delay as
\begin{small}
\begin{align}\label{eq:rendering}
  \mathbf{s}_{o}\left(\boldsymbol{\omega}_{o}, t\right) &= \int_{S^2} \int_{0}^{t} \cos \theta_i \mathbf{M}(\tau, \boldsymbol{\omega}_i, \boldsymbol{\omega}_o) \mathbf{s}_{i}(\boldsymbol{\omega}_i, t-\tau) \mathrm{d}\tau \mathrm{d}\boldsymbol{\omega}_i \nonumber\\
  &= \int_{S^2} \int_{0}^{t} \mathbf{H}(\tau, \boldsymbol{\omega}_i, \boldsymbol{\omega}_o) \mathbf{s}_{i}(\boldsymbol{\omega}_i, t-\tau) \mathrm{d}\tau \mathrm{d}\boldsymbol{\omega}_i,
\end{align}
\end{small}
where $\mathbf{H}$ is the Mueller matrix scaled by the cosine foreshortening.
As such, the Stokes vector of the outgoing light ($\mathbf{s}_i$) is the function of time and propagation direction.


\section{Polarimetric ToF Imaging}
\label{sec:imaging}
We now turn to developing a polarimetric ToF imaging system to capture temporal-polarimetric scene response.

\paragraph{Coaxial ToF Imaging}
Although the proposed method is not restricted to a specific time-resolved sensing system, we opt to use a pulsed laser illumination and a synchronized SPAD sensor in a coaxial configuration typical to LiDAR systems~\cite{schwarz2010mapping}. This means that the illumination and detection share the same optical path to optically sample direct illumination with minimal indirect illumination by interreflection. This allows us to simplify the rendering equation of Equation~\eqref{eq:rendering} by removing the outer integral over direction, that is
\begin{align}
\label{eq:image_formation_tof}
  I(t, \boldsymbol{\omega}) &= \left[\int_{0}^{t'}\mathbf{H}(\tau, \boldsymbol{\omega},\boldsymbol{\omega}) \mathbf{s}_\textrm{illum}(\boldsymbol{\omega},t'-\tau)\mathrm{d}\tau\right]_0 \nonumber \\
  &= \left[\mathbf{H}(t', \boldsymbol{\omega},\boldsymbol{\omega}) \mathbf{s}_\textrm{illum}(\boldsymbol{\omega},0)\right]_0,
\end{align}
where $I(t, \boldsymbol{\omega})$ is the captured intensity of light with ToF $t$ along the direction $\boldsymbol{\omega}$. Here, $c$ is the speed of light and $\mathbf{s}_\textrm{illum}$ is the Stokes vector of the laser illumination. $t'$ is the shifted time by the ToF ($t'=t-\frac{2d}{c}$), where $d$ is the Euclidean distance between the collocated laser at position $\mathbf{0}$, set to the origin without loss of generality, and a scene point $\mathbf{p}=\mathbf{0}+d\boldsymbol{\omega}$. In addition, the second integral over the temporal dimension is simplified as a result of using pulsed laser illumination.

\begin{figure}[t]
  \centering
  \includegraphics[width=\linewidth]{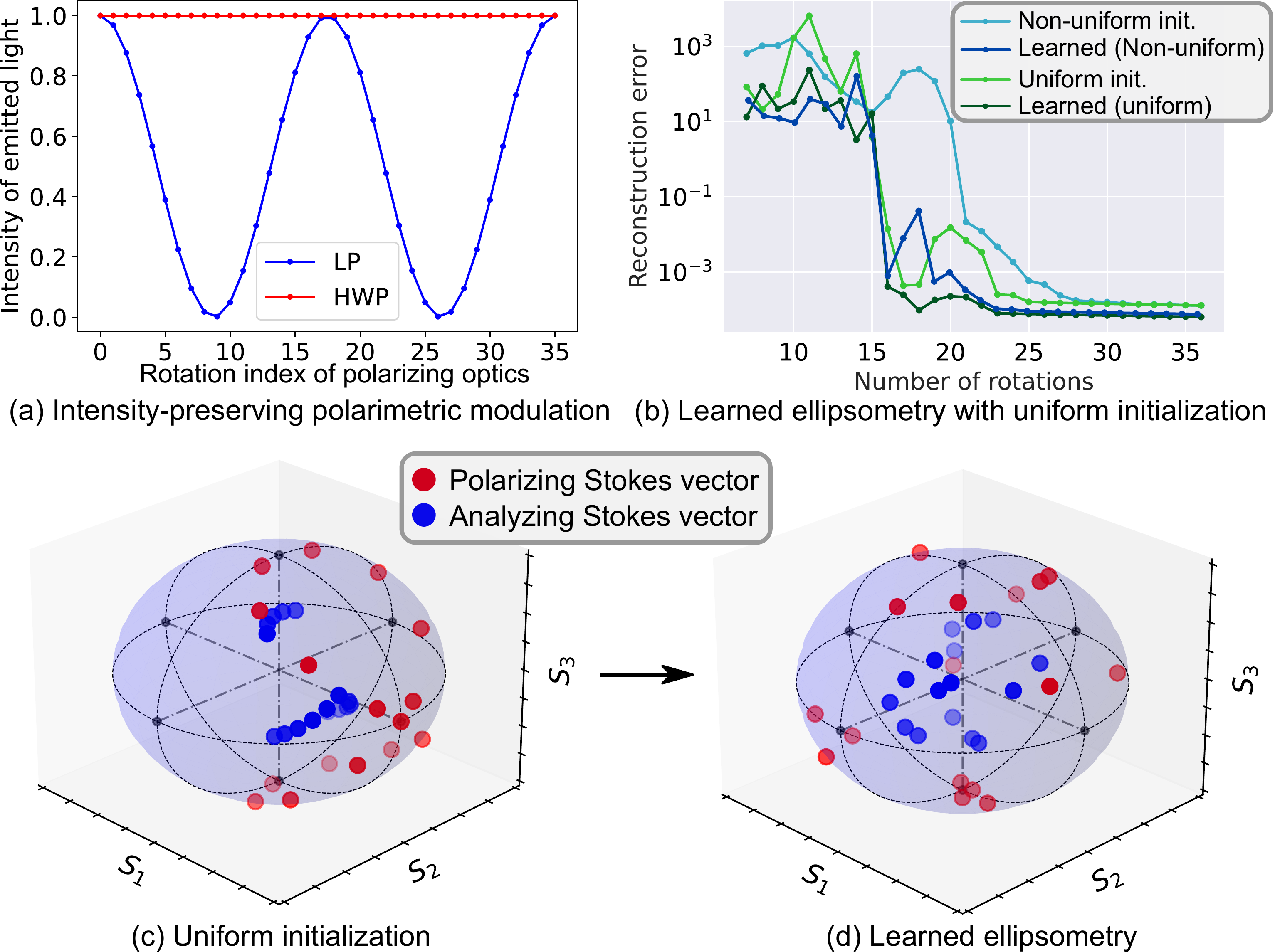}
  \caption{\label{fig:learned_ellipsometry}%
  Energy-efficient Polarimetric ToF Imaging. (a) We use a HWP instead of a LP to preserve the intensity of the polarized laser.
  We learn the rotation angles of the polarizing optics based on uniform initialization, resulting in (b) accurate reconstruction with fewer captures than conventional ellipsometry methods.
  We show the polarizing and analyzing Stokes vectors ($[\mathbf{P}_i]_{\forall,0}$, $[\mathbf{A}_i]_{0,\forall}$) on the Poincar\'{e} sphere before and after the learning.
  }
\end{figure}

\paragraph{Rotating Ellipsometry for Polarized Illumination}
Starting with the coaxial ToF imaging systems described above, we acquire the complete polarization change induced by light-matter interaction using rotating optical ellipsometry~\cite{collett2005field}. Rotating optical ellipsometry commonly incorporates pairs of linear polarizers (LP) and quarter-wave plates (QWP) in front of an \emph{unpolarized} illumination source and a detector~\cite{azzam1978photopolarimetric,baek2021probe}. Multiple intensity measurements $I_{i\in\{1,\cdots,N\}}$ are then captured by rotating the quarter-wave plates~\cite{azzam1978photopolarimetric}, or together with rotating linear polarizers~\cite{baek2021probe} to reduce the required capture counts.

Unfortunately, rotating linear polarizers in our ToF imaging setup results in \emph{significant} loss of light energy, because our laser illumination itself is linearly polarized; the polarized laser illumination loses energy as it passes through the linear polarizer. Instead of a linear polarizer, we present a simple solution for this light-inefficient polarization modulation by placing a half-wave plate (HWP) in front of the laser illumination.

This allows for rotating the polarization axis of the laser illumination without sacrificing illumination intensity as shown in Figure~\ref{fig:learned_ellipsometry}(a).

As a result, our coaxial polarimetric ToF imaging system consists of a pulsed laser, HWP, QWP, i.e., the illumination module, and QWP, LP, a SPAD forming an analyzer module.
The image formation of the proposed method is
\begin{equation}\label{eq:ellipsometry}
  I_i(t, \boldsymbol{\omega}) = \left[\mathbf{A}_i \mathbf{H}(t', \boldsymbol{\omega},\boldsymbol{\omega}) \mathbf{P}_i\mathbf{s}_\textrm{illum}(\boldsymbol{\omega},0)\right]_0,
\end{equation}
where $\mathbf{A}_i$ and $\mathbf{P}_i$ are the $i$-th Mueller matrices of the analyzing optics and the polarizing optics defined as
  $\mathbf{A}_i = \mathbf{L}(\theta_i^4)\mathbf{Q}(\theta_i^3)$ and $\mathbf{P}_i = \mathbf{Q}(\theta_i^2)\mathbf{W}(\theta_i^1),$
with $\theta_i^{\{1,2,3,4\}}$ as the rotation angles of the polarizing module HWP and QWP and the analyzing module QWP and LP.
Here, $\mathbf{W}$, $\mathbf{Q}$, and $\mathbf{L}$ are the Mueller matrices of HWP, QWP, and LP~\cite{collett2005field}.
Equation~\eqref{eq:ellipsometry} formulates the temporal-polarimetric transport in our system from laser illumination, polarizing optics, scene transport, analyzing optics, and SPAD detection.
The polarimetric response of the galvo-mirror is calibrated by capturing an uncoated gold mirror with known response~\cite{baek2020image}.

\paragraph{Learned Ellipsometry with Uniform Initialization}
Next, we investigate how few intensity measurements $I_i$ can be captured to faithfully reconstruct the Mueller matrix $\mathbf{H}$.
This is critical for the proposed polarimetric ToF imaging method, as it performs high-dimensional probing via sequential galvo scanning and rotation of polarization optics, resulting in hour-long capture time.
Recently, Baek and Heide~\cite{baek2021probe} presented a learned ellipsometry which optimizes the rotation angles of polarizing optics using first-order data-driven optimization.
In the same spirit, we learn the optimal rotation angles using a differentiable implementation of Equation~\eqref{eq:ellipsometry}.
One modification we make is to use a different initialization scheme for the angles, which is essential to overcome local minima.
Specifically, we initialize the rotation angles of the polarization optics to produce diverse polarization states, uniformly distributed the Poincar\'{e} sphere~\cite{collett2005field}.
Generated by the proposed initialization scheme, the experiments in Figure~\ref{fig:learned_ellipsometry}(b) validate that the polarization states provide more reliable constructions than using the non-uniform initialization~\cite{baek2021probe}.

%

\section{All-photon Scene Reconstruction}
\label{sec:shape_reconstruction}
The proposed temporal-polarimetric image formation model explains how geometric and material parameters of a scene relate to our sensor measurements in the presence of surface and sub-surface reflections. For the first time, this allows us to solve the corresponding inverse problem: Given the temporal-polarimetric sensor measurements $I_i$, we estimate a set of scene parameters $\boldsymbol{\Theta}$, including travel distance ($d$), surface normals ($\mathbf{n}$), refractive index ($\eta$), surface roughness ($m$), and scattering parameters ($a_i^{s/ss}, \mu_i^{s/ss}, \sigma_i^{s/ss}$): $\boldsymbol{\Theta}=\{d, \mathbf{n}, \eta, m, a_i^{s/ss}, \mu_i^{s/ss}, \sigma_i^{s/ss}\}$.
Note that conventional ToF imaging methods estimate travel distance by measuring the round-trip time of the first-returning photons with digital or analog peak-finding methods. In contrast, we do not discard photons that are arriving later and exploit the temporal-polarimetric information in the surface and sub-surface reflections to reconstruct the scene parameters.

\paragraph{Ellipsometric Reconstruction}
We first convert the temporal measurements $I_i$ for all the rotation angles of the polarizing optics into a time-varying Mueller matrix $\mathbf{H}_\mathrm{meas}$. To this end, we invert the image formation model of Equation~\eqref{eq:ellipsometry} in a least-squares manner following Baek et al.~\cite{baek2020image}, solving the following optimization problem for each spatio-temporal pixel
\begin{align}\label{eq:M_meas}
    \mathop {{\rm{minimize}}}\limits_{\mathbf{H}_\mathrm{meas}} \sum\limits_{i=1}^N
\left(I_i
- \left[ \mathbf{A}_i\,\mathbf{H}_\mathrm{meas}\, \mathbf{P}_i \mathbf{s}_\textrm{illum} \right]_{0}\right)^2.
\end{align}
We implement the corresponding solver efficiently by parallelizing the pseudo-inverse matrix-vector product, see details in the Supplemental Document.

\begin{figure}[t]
  \centering
  \includegraphics[width=\linewidth]{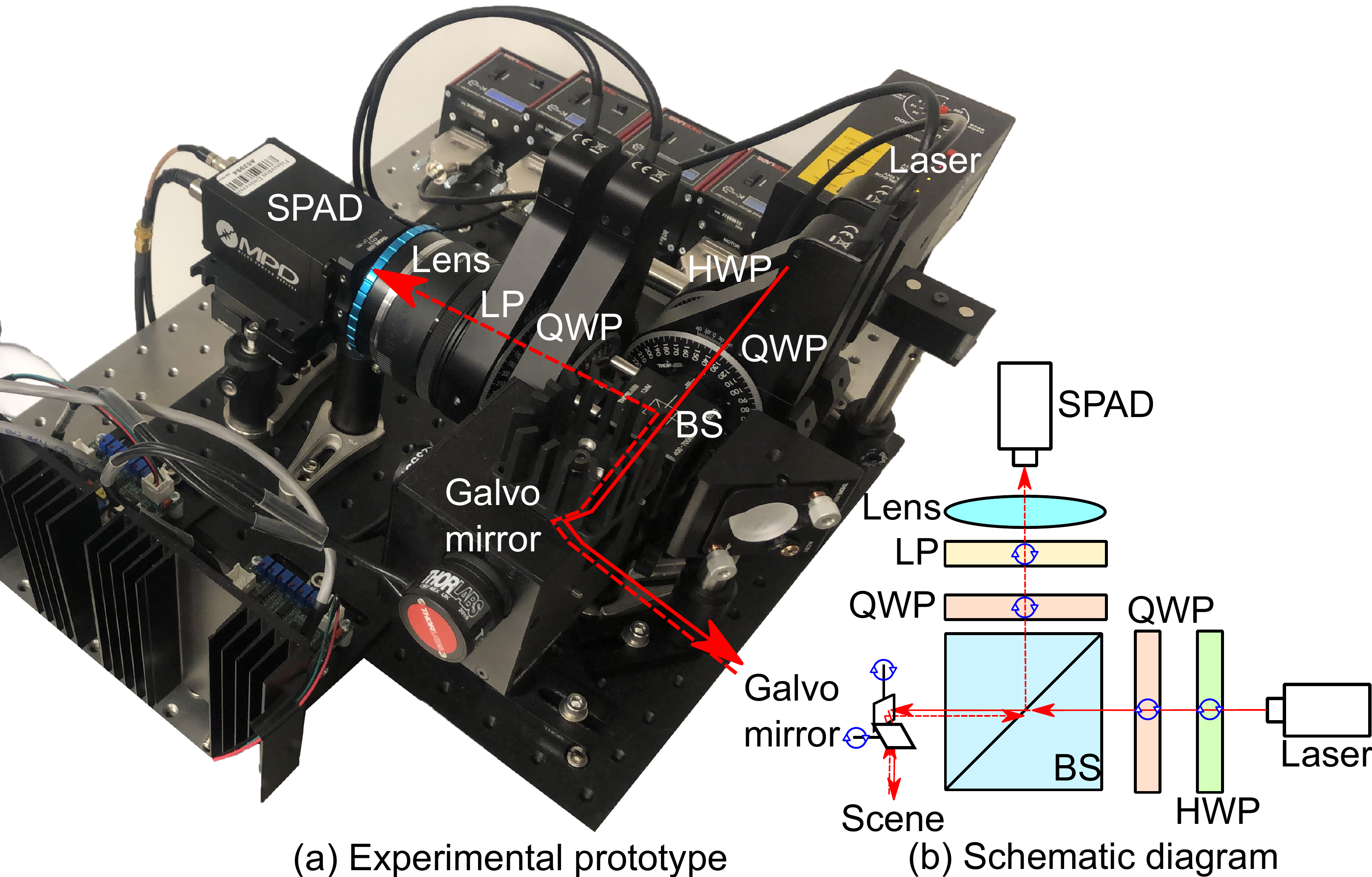}
  \caption{\label{fig:system}%
  Experimental Setup. We built (a) an experimental prototype for polarimetric ToF imaging, allowing us to efficiently capture temporal-polarimetric images. Its optical path is shown in (b).
  }
\end{figure}

\paragraph{Scene Reconstruction}
Once we have obtained the Mueller matrix $\mathbf{H}_\mathrm{meas}$ for each temporal bin $ t'$ and coaxial incident/outgoing direction $\boldsymbol{\omega}$, we estimate the scene parameters $\boldsymbol{\Theta}$ best explaining the Mueller matrix $\mathbf{H}_\mathrm{meas}$, that is
\begin{small}
\begin{align}\label{eq:scene_recon}
    \mathop {\rm{minimize}}\limits_{\dot{\boldsymbol{\Theta}}} \underbrace{\left\| \dot{\mathbf{W}}_p \odot \left( {f(\dot{\boldsymbol{{\Theta}}}) - {\dot{\mathbf{H}}_{{\mathrm{meas}}}}} \right) \right\|_1}_{\text{Data term}} + \underbrace{\left\| \dot{\mathbf{W}}_d \odot \nabla \dot{\mathbf{n}} \right\|_1}_{\text{Regularization term}},
\end{align}
\end{small}
where $\dot{\mathbf{x}}$ is the stacked vector of pixel quantities of variable $\mathbf{x}$ for each temporal bin, spatial direction, and Mueller matrix element. Here, $\odot$ is the Hadamard product, and $\nabla$ is the spatial gradient operator.
The data term evaluates the predicted Mueller matrix using the forward rendering function $f(\boldsymbol{\Theta})=\mathbf{H}(\tau,\boldsymbol{\omega},\boldsymbol{\omega};\boldsymbol{\Theta})$.
The matrix $\mathbf{W}_p$ weights the diagonal/non-diagonal Mueller matrix elements due to their different scales, see Supplemental Document for details.
The regularization term penalizes steep surface normal changes in the spatial dimension.
The matrix $\mathbf{W}_d$ is a weighting matrix that penalizes the the occlusion edges.
With the objective in Equation~\eqref{eq:scene_recon} being differentiable with respect to the scene parameters $\boldsymbol{\Theta}$, we solve it using the first-order Adam optimization~\cite{paszke2017automatic}.

To circumvent the ill-posedness of Equation~\eqref{eq:scene_recon}, we propose the three following modifications.
First, we take a regional approach commonly used in inverse rendering methods~\cite{meka2021real,baek2018simultaneous}.
That is, while surface normals and travel distance are per-pixel quantities, we define refractive index, surface roughness, and scattering parameters as per-material quantities specified by a given cluster map. This reduces the total parameter count.
We compute the cluster map based on the per-pixel average intensity using $k$-means clustering.
Second, we incorporate the unit-norm and range constraints of each scene parameter to reduce parameter search space with vector normalization and scaled sigmoid functions, see Supplemental Document for details.
Third, we initialize the travel distance $d$ and surface normals $\mathbf{n}$ with conventional peak-finding outputs on the unpolarized ToF measurements corresponding to the first element of the Mueller matrix $[\mathbf{H}_\mathrm{meas}]_{00}$.

\begin{figure}[t]
  \centering
  \includegraphics[width=\linewidth]{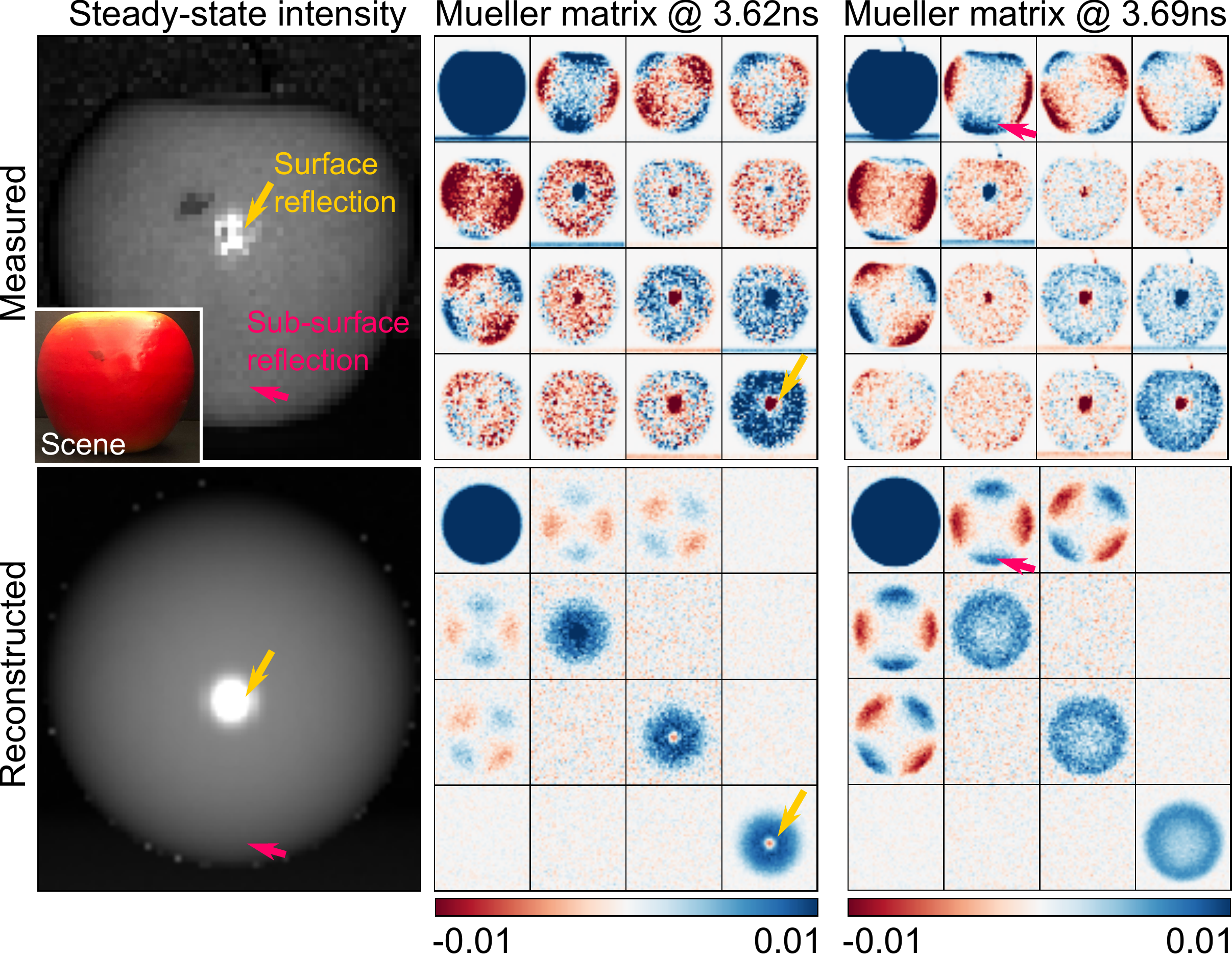}
  \caption{\label{fig:results_brdf}%
  Temporal-Polarimetric BRDF Analysis. The proposed BRDF model is the first parametric representation that is capable of describing both surface (orange arrows) and sub-surface reflection (pink arrows) by finding the optimal scene parameters of a spherical object. Our model predicts the normal-dependency of sub-surface reflection as well as the polarization-dependent sign changes of surface reflection.
  }
\end{figure}

\begin{figure*}[t]
  \centering
  \includegraphics[width=\linewidth]{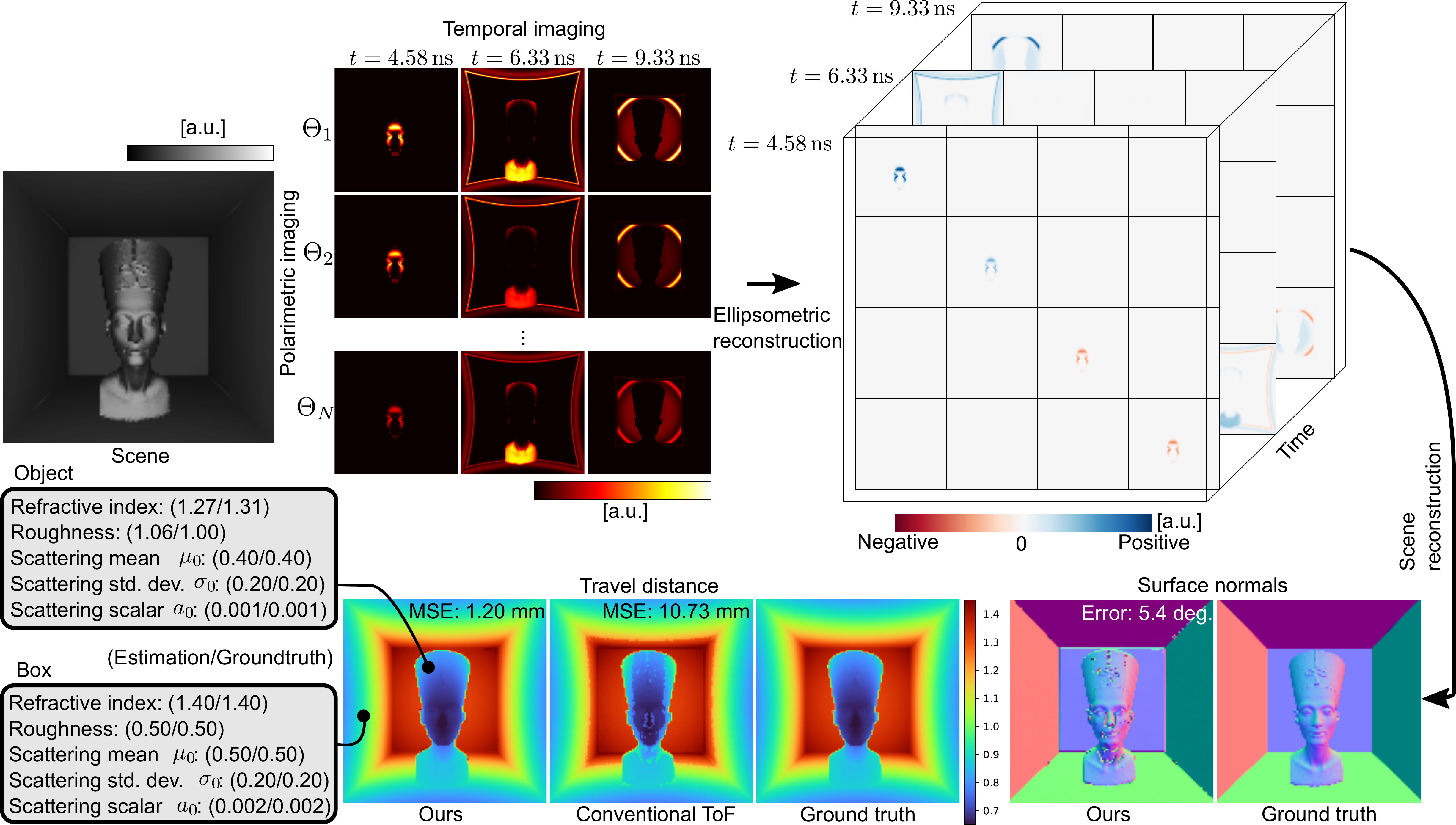}
  \caption{\label{fig:results_syn}%
  Synthetic Validation: We evaluate our all-photon ToF imaging on synthetic data simulating the entire process of temporal-polarimetric capture, ellipsometric reconstruction, and scene reconstruction. The proposed method accurately estimates travel distance, surface normals, and material parameters.
  }
\end{figure*}

\section{Experimental Setup}
Figure~\ref{fig:system} shows our experimental prototype and corresponding illustration.
We use a picosecond pulsed laser of optical wavelength 635\,nm (Edinburgh Instruments EPL-635) and a SPAD sensor (MPD Series) with 25\,ps temporal resolution. A time-correlation circuit (PicoQuant TimeHarp 260 PICO) synchronizes the laser and the SPAD. Light emitted from the laser passes through a half-wave plate (Thorlabs AHWP10M-600) and a quarter-wave plate (Thorlabs AQWP10M-580). A galvo mirror (Thorlabs GVS012) redirects the beam to a scene. We use a non-polarizing beamsplitter (Thorlabs BS013) to implement a coaxial configuration. Another quarter-wave plate and a linear polarizer (Newport 10LP-VIS-B) modulate the returning light. We use a 50\,mm objective lens for the SPAD sensor, and mount the polarizing optics on motorized rotary stages (Thorlabs KPRM1E). We implement capture and reconstruction scripts in Matlab and Pytorch, respectively.

\section{Assessment}
\label{sec:results}

\begin{figure*}[t]
  \centering
  \includegraphics[width=\linewidth]{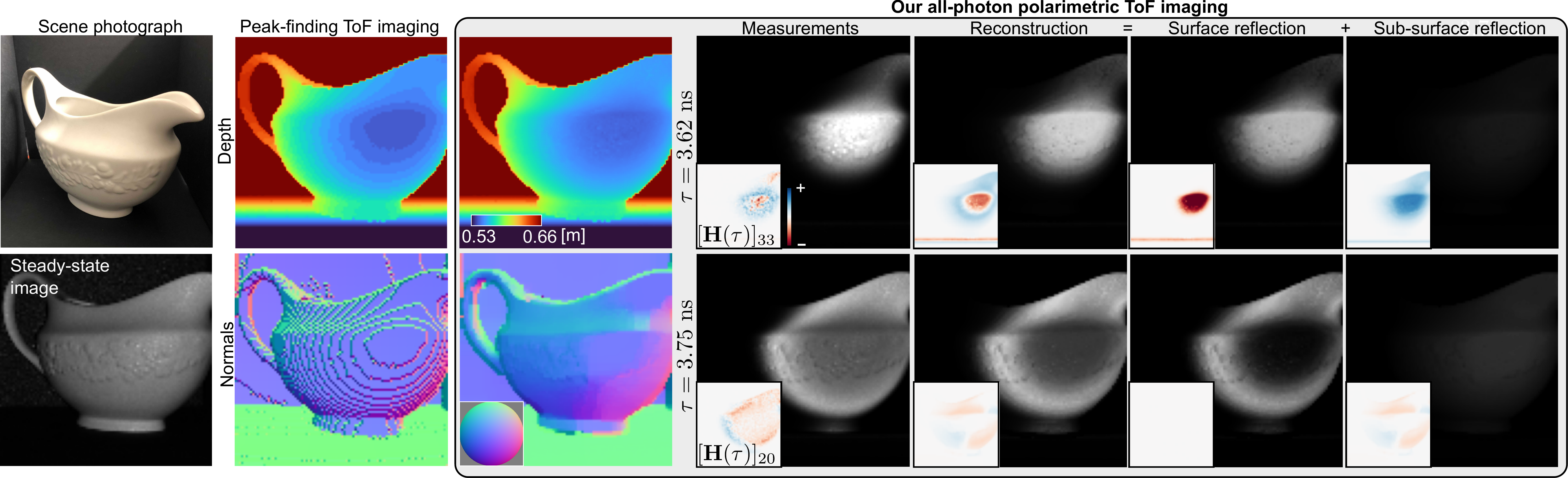}
  \caption{\label{fig:results_real}%
  Experimental Validation: We validate our all-photon polarimetric ToF imaging on real-world scenes. Conventional ToF imaging methods acquire scene depth from first-reflected photons only, resulting in inaccurate geometry reflected in the depth and surface normals estimates. In contrast, the proposed method relies on all photons, enabling accurate geometric reconstruction and the recovery of material parameters by decomposing light transport into surface and sub-surface components.
  }
\end{figure*}

\paragraph{Temporal-polarimetric BRDF}
To evaluate the representation capacity of our temporal-polarimetric BRDF model, we capture a spherical object and reconstruct its corresponding Mueller matrix for each time bin: $\mathbf{H}_\mathrm{meas}(t,\boldsymbol{\omega},\boldsymbol{\omega})$.
We then find the scene parameters $\mathbf{\Theta}$ of a spherical object, which best explains the observed temporal Mueller matrices, resulting in the reconstructed Mueller matrix $\mathbf{H}(t,\boldsymbol{\omega},\boldsymbol{\omega};\mathbf{\Theta})$.
We further convert $\mathbf{H}$ to another Mueller matrix $\mathbf{H}'$ so that it follows the similar noise statistics of the measured Mueller matrix $\mathbf{H}_\mathrm{meas}$ by simulating the sensor measurements $I'_i$ with the reconstructed Mueller matrix $\mathbf{H}$ and again reconstructing this Mueller matrix.
Figure~\ref{fig:results_brdf} validates that the proposed BRDF model accurately predicts the temporal-polarimetric reflection of the spherical object.
In particular, our model accurately predicts the normal-dependent temporal-polarimetric structure of sub-surface reflection at the boundaries of the spherical object in $\mathbf{H}'_{\{i,0\}, \{0,i\}, \forall i}$. 
We also observe the accurate sign changes of the circular polarimetric sub-surface and surface reflection in $\mathbf{H}'_{3,3}$.


\paragraph{Synthetic Evaluation}
We evaluate the proposed method on three synthetic scenes shown in Figure~\ref{fig:results_syn}.
As ground-truth scene parameters $\mathbf{\Theta}$ for the synthetic scenes are known, we simulate the synthetic sensor measurements $I_i$ using the image formation model from Equation~\eqref{eq:ellipsometry}.
Here, we add Gaussian noise with a standard deviation of $10^{-4}$ to simulate measurement noise. 
With the simulated observations in hand, we solve Equation~\eqref{eq:scene_recon} by performing the proposed all-photon scene reconstruction, that is ellipsometric reconstruction of the temporal-polarimetric Mueller matrix $\mathbf{H}_\mathrm{meas}$ and scene reconstruction.
Figure~\ref{fig:results_syn} validates that our reconstruction approach estimates geometric and material parameters accurately, both qualitatively and quantitatively.
For additional results, we refer to the Supplemental Document.

\begin{figure}[t]
  \centering
  \includegraphics[width=\linewidth]{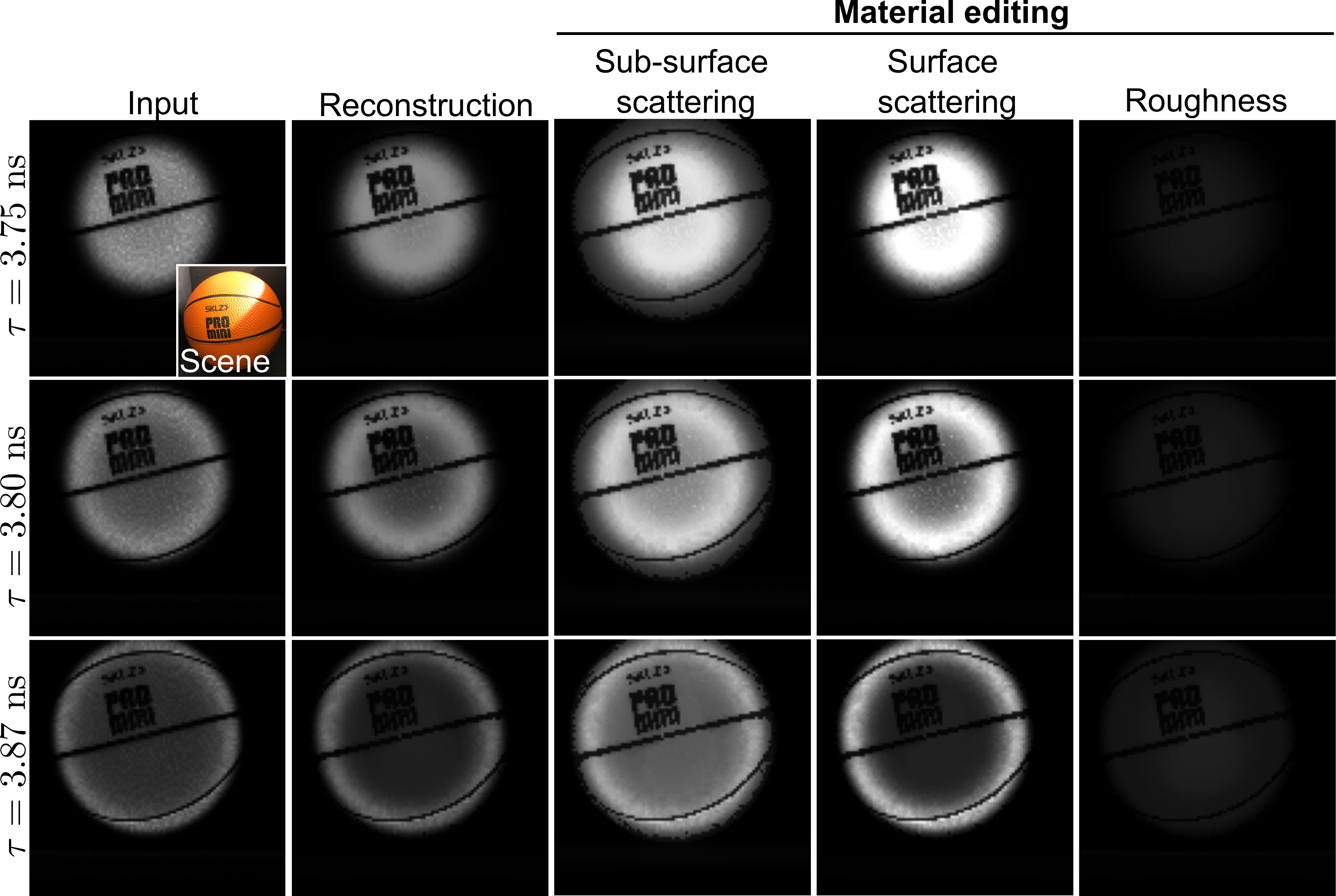}
  \caption{\label{fig:results_real_material}%
  Material Editing. We edit the material parameters reconstructed from our polarimetric ToF imaging. Specifically, we change the scattering parameters of sub-surface and surface reflection as well as the surface roughness. We refer to the text for details.
  }
\end{figure}

\paragraph{Experimental Evaluation}
Next, we validate the proposed method on the additional real-world scenes shown in Figures~\ref{fig:results_real} and ~\ref{fig:results_real_material}.
We reconstruct the temporal-polarimetric Mueller matrix $\mathbf{H}_\mathrm{meas}$ from experimental measurements $I_i$ followed by finding the optimal scene parameters $\mathbf{\Theta}$.
Our all-photon polarimetric ToF imaging reconstructs accurate travel distance estimates that correspond to per-point depth and surface normals. 
We also acquire scattering parameters of the scene. Together with the obtained geometry, the proposed method accurately reconstructs the original measurements.
Figure~\ref{fig:results_real} shows the rendered measurements with our estimated parameters $\mathbf{\Theta}$ and the corresponding sensor measurements.
In addition, our scene reconstructions faciliate for material editing, which we demonstrate in three editing examples in each column of Figure~\ref{fig:results_real_material}.
We first change the sub-surface scattering parameters by increasing its scalar by three times as $a_i \leftarrow 3a_i$, and changing its polarization-dependent temporal shift as $\boldsymbol{\mu} \leftarrow [0.1\mu_0, 2\mu_1, 2\mu_2, 2\mu_3]$.
This results in stronger sub-surface scattering. 
We then edit the surface scattering parameters with the same amount for the shift, but doubling the scalar $a_i \leftarrow 2a_i$.
This mimics a material with stronger surface reflection. 
Lastly, increasing the surface roughness provides the appearance of rough surface with reflection dominated by sub-surface reflection.

\section{Conclusion}
\label{sec:conclusion}
We introduce an all-photon polarimetric ToF imaging method, which extracts information from both surface and sub-surface reflection for acquiring geometry and scattering parameters. To this end, we propose the first temporal-polarimetric BRDF model, which relates the sensor measurements to geometric and scattering parameters of a scene. The proposed learned ellipsometric ToF imaging system permits the acquisition of temporal-polarimetric data with energy-efficient polarimetric modulation. From this, we reconstruct scene parameters and rely on all photons instead of the first reflected ones. We validate the method in simulation and experimentally, asserting that the method outperforms existing ToF imaging methods. While we validate the utility of temporal-polarimetric measurements, our experimental setup is restricted to sequential scanning by the mechanical rotation of the polarization optics available to us. Exciting future directions include equipping the proposed system with electronically-controllable liquid crystal modulators, or combining SPAD array sensors with polarization filter arrays, which may pave the way towards single-shot temporal-polarimetric acquisition. Aside from making a step towards high-dimensional computational imaging, the approach could also pave the way for practical implementation such as gated imaging with polarization cues as a low-cost automotive imaging modality. 

{\small
\bibliographystyle{ieee_fullname}
\bibliography{egbib}

\begin{thebibliography}{10}\itemsep=-1pt

\bibitem{abdalati2010icesat}
Waleed Abdalati, H~Jay Zwally, Robert Bindschadler, Bea Csatho, Sinead~Louise
  Farrell, Helen~Amanda Fricker, David Harding, Ronald Kwok, Michael Lefsky,
  Thorsten Markus, et~al.
\newblock The icesat-2 laser altimetry mission.
\newblock {\em Proceedings of the IEEE}, 98(5):735--751, 2010.

\bibitem{atkinson2006recovery}
Gary~A Atkinson and Edwin~R Hancock.
\newblock Recovery of surface orientation from diffuse polarization.
\newblock {\em IEEE Transactions on Image Processing (TIP)}, 15(6):1653--1664,
  2006.

\bibitem{azzam1978photopolarimetric}
RMA Azzam.
\newblock Photopolarimetric measurement of the mueller matrix by fourier
  analysis of a single detected signal.
\newblock {\em Optics Letters}, 2(6):148--150, 1978.

\bibitem{baek2016birefractive}
Seung-Hwan Baek, Diego Gutierrez, and Min~H Kim.
\newblock Birefractive stereo imaging for single-shot depth acquisition.
\newblock {\em ACM Transactions on Graphics (TOG)}, 35(6):1--11, 2016.

\bibitem{baek2021probe}
Seung-Hwan Baek and Felix Heide.
\newblock Polarimetric spatio-temporal light transport probing.
\newblock {\em ACM Transactions on Graphics (TOG)}, 2021.

\bibitem{baek2018simultaneous}
Seung-Hwan Baek, Daniel~S Jeon, Xin Tong, and Min~H Kim.
\newblock Simultaneous acquisition of polarimetric svbrdf and normals.
\newblock {\em ACM Transactions on Graphics (TOG)}, 37(6):268--1, 2018.

\bibitem{baek2020image}
Seung-Hwan Baek, Tizian Zeltner, Hyunjin Ku, Inseung Hwang, Xin Tong, Wenzel
  Jakob, and Min~H Kim.
\newblock Image-based acquisition and modeling of polarimetric reflectance.
\newblock {\em ACM Transactions on Graphics (TOG)}, 39(4):139, 2020.

\bibitem{callenberg2017snapshot}
Clara Callenberg, Felix Heide, Gordon Wetzstein, and Matthias Hullin.
\newblock Snapshot difference imaging using time-of-flight sensors.
\newblock {\em ACM Transactions on Graphics (TOG)}, 2017.

\bibitem{chang2017matterport3d}
Angel Chang, Angela Dai, Thomas Funkhouser, Maciej Halber, Matthias Niessner,
  Manolis Savva, Shuran Song, Andy Zeng, and Yinda Zhang.
\newblock Matterport3d: Learning from rgb-d data in indoor environments.
\newblock {\em International Conference on 3D Vision (3DV)}, 2017.

\bibitem{collett2005field}
Edward Collett.
\newblock Field guide to polarization.
\newblock Spie Bellingham, WA, 2005.

\bibitem{dai2017scannet}
Angela Dai, Angel~X Chang, Manolis Savva, Maciej Halber, Thomas Funkhouser, and
  Matthias Nie{\ss}ner.
\newblock Scannet: Richly-annotated 3d reconstructions of indoor scenes.
\newblock In {\em IEEE Conference on Computer Vision and Pattern Recognition
  (CVPR)}, pages 5828--5839, 2017.

\bibitem{germer2020evolution}
Thomas~A Germer.
\newblock Evolution of transmitted depolarization in diffusely scattering
  media.
\newblock {\em Journal of the Optical Society of America (JOSA)},
  37(6):980--987, 2020.

\bibitem{ghosh2010circularly}
Abhijeet Ghosh, Tongbo Chen, Pieter Peers, Cyrus~A Wilson, and Paul Debevec.
\newblock Circularly polarized spherical illumination reflectometry.
\newblock pages 1--12, 2010.

\bibitem{goldstein2017polarized}
Dennis~H Goldstein.
\newblock {\em Polarized light}.
\newblock CRC press, 2017.

\bibitem{heide2018sub}
Felix Heide, Steven Diamond, David~B Lindell, and Gordon Wetzstein.
\newblock Sub-picosecond photon-efficient 3d imaging using single-photon
  sensors.
\newblock {\em Scientific Reports}, 8(1):1--8, 2018.

\bibitem{heitz2014understanding}
Eric Heitz.
\newblock Understanding the masking-shadowing function in microfacet-based
  brdfs.
\newblock {\em Journal of Computer Graphics Techniques}, 3(2):32--91, 2014.

\bibitem{huffman2020real}
J~Alex Huffman, Anne~E Perring, Nicole~J Savage, Bernard Clot, Beno{\^\i}t
  Crouzy, Fiona Tummon, Ofir Shoshanim, Brian Damit, Johannes Schneider,
  Vasanthi Sivaprakasam, et~al.
\newblock Real-time sensing of bioaerosols: Review and current perspectives.
\newblock {\em Aerosol Science and Technology}, 54(5):465--495, 2020.

\bibitem{hyde2009geometrical}
Milo~W Hyde~IV, Jason~D Schmidt, and Michael~J Havrilla.
\newblock A geometrical optics polarimetric bidirectional reflectance
  distribution function for dielectric and metallic surfaces.
\newblock {\em Optics Express}, 17(24):22138--22153, 2009.

\bibitem{kadambi2015polarized}
Achuta Kadambi, Vage Taamazyan, Boxin Shi, and Ramesh Raskar.
\newblock Polarized 3d: High-quality depth sensing with polarization cues.
\newblock In {\em IEEE International Conference on Computer Vision (ICCV)},
  pages 3370--3378, 2015.

\bibitem{kajiya1986rendering}
James~T Kajiya.
\newblock The rendering equation.
\newblock In {\em Proceedings of the 13th Annual Conference on Computer
  Graphics and Interactive Techniques}, pages 143--150, 1986.

\bibitem{kim2021nanophotonics}
Inki Kim, Renato~Juliano Martins, Jaehyuck Jang, Trevon Badloe, Samira Khadir,
  Ho-Youl Jung, Hyeongdo Kim, Jongun Kim, Patrice Genevet, and Junsuk Rho.
\newblock Nanophotonics for light detection and ranging technology.
\newblock {\em Nature Nanotechnology}, 16(5):508--524, 2021.

\bibitem{lange2001solid}
Robert Lange and Peter Seitz.
\newblock Solid-state time-of-flight range camera.
\newblock {\em Journal of Quantum Electronics}, 37(3):390--397, 2001.

\bibitem{levin2007image}
Anat Levin, Rob Fergus, Fr{\'e}do Durand, and William~T Freeman.
\newblock Image and depth from a conventional camera with a coded aperture.
\newblock {\em ACM Transactions on Graphics (TOG)}, 26(3):70--es, 2007.

\bibitem{lindell2018single}
David~B Lindell, Matthew O'Toole, and Gordon Wetzstein.
\newblock Single-photon 3d imaging with deep sensor fusion.
\newblock {\em ACM Transactions on Graphics (TOG)}, 37(4):113--1, 2018.

\bibitem{ma2007rapid}
Wan-Chun Ma, Tim Hawkins, Pieter Peers, Charles-Felix Chabert, Malte Weiss,
  Paul~E Debevec, et~al.
\newblock Rapid acquisition of specular and diffuse normal maps from polarized
  spherical gradient illumination.
\newblock {\em Eurographics Symposium on Rendering (EGSR)}, 2007(9):10, 2007.

\bibitem{mccarthy2009long}
Aongus McCarthy, Robert~J Collins, Nils~J Krichel, Ver{\'o}nica Fern{\'a}ndez,
  Andrew~M Wallace, and Gerald~S Buller.
\newblock Long-range time-of-flight scanning sensor based on high-speed
  time-correlated single-photon counting.
\newblock {\em Applied Optics}, 48(32):6241--6251, 2009.

\bibitem{meka2021real}
Abhimitra Meka, Mohammad Shafiei, Michael Zollh{\"o}fer, Christian Richardt,
  and Christian Theobalt.
\newblock Real-time global illumination decomposition of videos.
\newblock {\em ACM Transactions on Graphics (TOG)}, 40(3):1--16, 2021.

\bibitem{nayar1997separation}
Shree~K Nayar, Xi-Sheng Fang, and Terrance Boult.
\newblock Separation of reflection components using color and polarization.
\newblock {\em Springer International Journal of Computer Vision (IJCV)},
  21(3):163--186, 1997.

\bibitem{oren1994generalization}
Michael Oren and Shree~K Nayar.
\newblock Generalization of lambert's reflectance model.
\newblock In {\em Proceedings of the 21st Annual Conference on Computer
  Graphics and Interactive Techniques}, pages 239--246, 1994.

\bibitem{paszke2017automatic}
Adam Paszke, Sam Gross, Soumith Chintala, Gregory Chanan, Edward Yang, Zachary
  DeVito, Zeming Lin, Alban Desmaison, Luca Antiga, and Adam Lerer.
\newblock Automatic differentiation in pytorch.
\newblock 2017.

\bibitem{rapp2017few}
Joshua Rapp and Vivek~K Goyal.
\newblock A few photons among many: Unmixing signal and noise for
  photon-efficient active imaging.
\newblock {\em IEEE Transactions on Computational Imaging (TCI)},
  3(3):445--459, 2017.

\bibitem{renker2006geiger}
Dieter Renker.
\newblock Geiger-mode avalanche photodiodes, history, properties and problems.
\newblock {\em Nuclear Instruments and Methods in Physics Research Section A:
  Accelerators, Spectrometers, Detectors and Associated Equipment},
  567(1):48--56, 2006.

\bibitem{sassen1991polarization}
Kenneth Sassen.
\newblock The polarization lidar technique for cloud research: A review and
  current assessment.
\newblock {\em Bulletin of the American Meteorological Society},
  72(12):1848--1866, 1991.

\bibitem{sassen2005polarization}
Kenneth Sassen.
\newblock Polarization in lidar.
\newblock In {\em Lidar}, pages 19--42. Springer, 2005.

\bibitem{satat2016all}
Guy Satat, Barmak Heshmat, Dan Raviv, and Ramesh Raskar.
\newblock All photons imaging through volumetric scattering.
\newblock {\em Scientific Reports}, 6(1):1--8, 2016.

\bibitem{satat2018towards}
Guy Satat, Matthew Tancik, and Ramesh Raskar.
\newblock Towards photography through realistic fog.
\newblock In {\em Int. Conf. Comput. Photog. (ICCP)}, pages 1--10. IEEE, 2018.

\bibitem{scharstein2002taxonomy}
Daniel Scharstein and Richard Szeliski.
\newblock A taxonomy and evaluation of dense two-frame stereo correspondence
  algorithms.
\newblock {\em Springer International Journal of Computer Vision (IJCV)},
  47(1):7--42, 2002.

\bibitem{schechner2001instant}
Yoav~Y Schechner, Srinivasa~G Narasimhan, and Shree~K Nayar.
\newblock Instant dehazing of images using polarization.
\newblock In {\em IEEE Conference on Computer Vision and Pattern Recognition
  (CVPR)}, volume~1, pages I--I. IEEE, 2001.

\bibitem{schotland1971observations}
Richard~M Schotland, Kenneth Sassen, and Richard Stone.
\newblock Observations by lidar of linear depolarization ratios for
  hydrometeors.
\newblock {\em Journal of Applied Meteorology and Climatology},
  10(5):1011--1017, 1971.

\bibitem{schwarz2010mapping}
Brent Schwarz.
\newblock Mapping the world in 3d.
\newblock {\em Nature Photonics}, 4(7):429--430, 2010.

\bibitem{shin2015photon}
Dongeek Shin, Ahmed Kirmani, Vivek~K Goyal, and Jeffrey~H Shapiro.
\newblock Photon-efficient computational 3-d and reflectivity imaging with
  single-photon detectors.
\newblock {\em IEEE Transactions on Computational Imaging (TCI)},
  1(2):112--125, 2015.

\bibitem{shin2016photon}
Dongeek Shin, Feihu Xu, Dheera Venkatraman, Rudi Lussana, Federica Villa,
  Franco Zappa, Vivek~K Goyal, Franco~NC Wong, and Jeffrey~H Shapiro.
\newblock Photon-efficient imaging with a single-photon camera.
\newblock {\em Nature Communications}, 7(1):1--8, 2016.

\bibitem{replica19arxiv}
Julian Straub, Thomas Whelan, Lingni Ma, Yufan Chen, Erik Wijmans, Simon Green,
  Jakob~J. Engel, Raul Mur-Artal, Carl Ren, Shobhit Verma, Anton Clarkson,
  Mingfei Yan, Brian Budge, Yajie Yan, Xiaqing Pan, June Yon, Yuyang Zou,
  Kimberly Leon, Nigel Carter, Jesus Briales, Tyler Gillingham, Elias Mueggler,
  Luis Pesqueira, Manolis Savva, Dhruv Batra, Hauke~M. Strasdat, Renzo~De
  Nardi, Michael Goesele, Steven Lovegrove, and Richard Newcombe.
\newblock The {R}eplica dataset: A digital replica of indoor spaces.
\newblock {\em arXiv preprint arXiv:1906.05797}, 2019.

\bibitem{torrance1967theory}
Kenneth~E Torrance and Ephraim~M Sparrow.
\newblock Theory for off-specular reflection from roughened surfaces.
\newblock {\em Journal of the Optical Society of America (JOSA)},
  57(9):1105--1114, 1967.

\bibitem{treibitz2008active}
Tali Treibitz and Yoav~Y Schechner.
\newblock Active polarization descattering.
\newblock {\em IEEE Transactions on Pattern Analysis and Machine Intelligence
  (TPAMI)}, 31(3):385--399, 2008.

\bibitem{vasilkov2001airborne}
Alexander~P Vasilkov, Yury~A Goldin, Boris~A Gureev, Frank~E Hoge, Robert~N
  Swift, and C~Wayne Wright.
\newblock Airborne polarized lidar detection of scattering layers in the ocean.
\newblock {\em Applied Optics}, 40(24):4353--4364, 2001.

\bibitem{walter2007microfacet}
Bruce Walter, Stephen~R Marschner, Hongsong Li, and Kenneth~E Torrance.
\newblock Microfacet models for refraction through rough surfaces.
\newblock {\em Rendering techniques}, 2007:18th, 2007.

\bibitem{winker2009overview}
David~M Winker, Mark~A Vaughan, Ali Omar, Yongxiang Hu, Kathleen~A Powell,
  Zhaoyan Liu, William~H Hunt, and Stuart~A Young.
\newblock Overview of the calipso mission and caliop data processing
  algorithms.
\newblock {\em Journal of Atmospheric and Oceanic Technology},
  26(11):2310--2323, 2009.

\bibitem{wu2014decomposing}
Di Wu, Andreas Velten, Matthew O’Toole, Belen Masia, Amit Agrawal, Qionghai
  Dai, and Ramesh Raskar.
\newblock Decomposing global light transport using time of flight imaging.
\newblock {\em Springer International Journal of Computer Vision (IJCV)},
  107(2):123--138, 2014.

\bibitem{zhang2012microsoft}
Zhengyou Zhang.
\newblock Microsoft kinect sensor and its effect.
\newblock {\em IEEE Multimedia}, 19(2):4--10, 2012.

\end{thebibliography}
}

\end{document}